\documentclass{article}

\usepackage{arxiv}

\usepackage[utf8]{inputenc} 
\usepackage[T1]{fontenc}    
\usepackage{hyperref}       
\usepackage{url}            
\usepackage{booktabs}       
\usepackage{amsfonts}       
\usepackage{nicefrac}       
\usepackage{microtype}      
\usepackage[title]{appendix}
\usepackage{graphicx}
\usepackage[bf]{caption}

\usepackage{array}

\usepackage[linesnumbered,ruled]{algorithm2e}
\usepackage{algpseudocode}
\usepackage{amsmath}
\usepackage{xcolor}

\SetCommentSty{mycommfont}
\newlength\algowd
\def\savewd#1{\setbox0=\hbox{#1\hspace{.7in}}\algowd=\wd0\relax#1}
\newcommand\algolines[2]{\savewd{#1}%
  \tcp*{\parbox[t]{\dimexpr\algowidth-\algowd\textcolor{blue}}{#2}}}

\usepackage{textcomp}

\usepackage{doi}
\usepackage{etoolbox}
\usepackage[natbibapa]{apacite}
\AtBeginDocument{}
\renewcommand{\APACrefnote}[1]{}
\newtoggle{bibdoi}
\newtoggle{biburl}
\makeatletter
\newsavebox{\bib@url}
\newsavebox{\bib@doi}

\undef{\APACrefURL}
\undef{\endAPACrefURL}
\undef{\APACrefDOI}
\undef{\endAPACrefDOI}

\newenvironment{APACrefURL}
  {\global\toggletrue{biburl}\lrbox\bib@url}
  {\endlrbox}

\newenvironment{APACrefDOI}
  {\global\toggletrue{bibdoi}\lrbox\bib@doi}
  {\endlrbox}

\newcommand{\printinfo}{
  \iftoggle{bibdoi}{\usebox{\bib@doi}}{\usebox{\bib@url}}
  \togglefalse{bibdoi}
}

\AtBeginEnvironment{thebibliography}{
\pretocmd{\PrintBackRefs}{%
  \iftoggle{bibdoi}
    {\iftoggle{biburl}{\unskip\unskip}{}\usebox{\bib@doi}}
    {\iftoggle{biburl}{Retrieved from \usebox{\bib@url}}}{}
  \togglefalse{bibdoi}\togglefalse{biburl}%
}{}{}}
\makeatother

\title{ELEV-VISION-SAM: Integrated Vision Language and Foundation Model for Automated Estimation of Building Lowest Floor Elevation }

\date{} 					

\renewcommand{\headeright}{}
\renewcommand{\undertitle}{}
\renewcommand{\shorttitle}{}

\hypersetup{
pdftitle={},
pdfsubject={},
}

\begin{document}
\maketitle

\begin{center}
{\Large
Yu-Hsuan Ho\textsuperscript{a},
Longxiang Li\textsuperscript{b},
Ali Mostafavi\textsuperscript{a,*}
\par}

\bigskip
\textsuperscript{a} Urban Resilience.AI Lab, Zachry Department of Civil and Environmental Engineering,\\ Texas A\&M University, College Station, TX\\
\textsuperscript{b} Department of Computer Science and Engineering,\\ Texas A\&M University, TX\\
\vspace{6pt}
\textsuperscript{*} corresponding author, email: amostafavi@civil.tamu.edu
\\
\end{center}
\bigskip
\begin{abstract}
Street view imagery, aided by advancements in image quality and accessibility, has emerged as a valuable resource for urban analytics research. Recent studies have explored its potential for estimating lowest floor elevation (LFE), offering a scalable alternative to traditional on-site  measurements, crucial for assessing properties' flood risk and damage extent. While existing methods rely on object detection, the introduction of image segmentation has broadened street view images' utility for LFE estimation, although challenges still remain in segmentation quality and capability to distinguish front doors from other doors. To address these challenges in LFE estimation, this study integrates the Segment Anything model, a segmentation foundation model, with vision language models to conduct text-prompt image segmentation on street view images for LFE estimation. By evaluating various vision language models, integration methods, and text prompts, we identify the most suitable model for street view image analytics and LFE estimation tasks, thereby improving the availability of the current LFE estimation model based on image segmentation from 33\% to 56\% of properties. Remarkably, our proposed method significantly enhances the availability of LFE estimation to almost all properties in which the front door is visible in the street view image. Also the findings present the first baseline and comparison of various vision models of street view image-based LFE estimation. The model and findings not only contribute to advancing street view image segmentation for urban analytics but also provide a novel approach for image segmentation tasks for other civil engineering and infrastructure analytics tasks.

\end{abstract}



\section{Introduction}
\label{sec:1}
Driven by climate change, both the frequency and intensity of floods increasing, particularly in the Northeast and South Central regions of the United States, with the number of ``Billion Dollar Disasters" increasing from one every two years in the 1980s to around 10 per year since 2010 \citep{cigler2017us}, making floods the most costly natural disasters in terms of finance and people affected \citep{stromberg2007natural,yin2023integrated,kousky2018financing}. Particularly in urban areas, flooding  has significant ramifications across social and economic domains \citep{yin2023unsupervised}. Precisely evaluating property flood risk and estimating potential damage are essential for implementing effective measures aimed at responding to flooding events and mitigating associated hazards \citep{liu2024floodgenome,ma2024urban}. One essential metric for assessing property flood risk and anticipating the extent of flood damage to buildings is the lowest floor elevation (LFE) of a building \citep{bodoque2016flood, zarekarizi2020neglecting, gao_exploring_2023}. The lowest floor of a building refers to the lowest floor of the lowest enclosed area, including a basement but excluding enclosures used for parking, building access, storage, or flood-resistance \citep{fema_index}. The LFE represents the measured height of a building's lowest floor relative to the National Geodetic Vertical Datum (NGVD) or another specified datum indicated on the Flood Insurance Rate Map (FIRM) for the corresponding area \citep{fema_index}. The traditional LFE measuring method is on-site manual inspection using a total station theodolite, a process that incurs significant costs in terms of time, finances, and human resources. Prior research has explored the utilization of LiDAR (light detection and ranging) data to accelerate LFE measurement. For example, \citet{xia2024computer} used LiDAR systems on vehicular platforms to collect LiDAR point cloud data for extracting LFE information. Mobile LiDAR data, however, remains expensive. A more accessible alternative is warranted.

Street view imagery (SVI) is emerging as a vital data source for urban analytics, driven by advancements in image quality and accessibility \citep{biljecki2021street,ibrahim2020understanding,kang2020review}. Recent studies have explored its potential for estimating LFE as a more scalable alternative to conventional in-situ measurements \citep{ho2023elevvision, ning_exploring_2022, gao_exploring_2023}. Initial studies in this area proposed object detection techniques for LFE estimation using street view images, as demonstrated by \citet{ning_exploring_2022}, who utilized this approach to identify door bottoms in re-projected perspective Google Street View images. Subsequently, image segmentation techniques were introduced to implement LFE estimation directly upon panoramic street view images and expand the utility to various flood-related building elevation information. \citet{ho2023elevvision} proposed ELEV-VISION, which uses instance segmentation and semantic segmentation on panoramic street view images to extract the edges of front doors and roadsides for estimating LFE and the height difference between the street and the lowest floor (HDSL). These initial studies have shown the potential of image segmentation for building lowest flood elevation estimation; however, important limitations still persist.

Image segmentation on street view images is also essential for urban environment assessment and urban transportation analysis, with numerous downstream applications focusing on scene composition analysis. \citet{sanchez2024accessing} employed semantic segmentation on street view images to assess greenness visibility. \citet{fei2024adapting} integrated semantic information from street view images with a land-use regression model to enhance traffic-related air pollution estimation. \citet{narazaki2020vision} applied semantic segmentation for bridge component recognition on multiple image sources, including street view images. Compared to most of the tasks in the urban analytics field, vertical information extraction requires fine-grained segmentation and high-quality masks due to the utilization of segmentation outputs in computing vertical information. \citet{lu2023automated} estimated vehicle heights by constructing 3D bounding boxes based on image segmentation and selecting the object in the traffic scene with a known height as the reference. Unlike fixed-scene height estimation tasks, the reference height is less reliable in building vertical information extraction. \citet{xu2023building} calculated building heights by employing instance segmentation on panoramic street view images, using multiple images to enhance calculation. Combining information from street view images in multiple viewpoints is essential for building information extraction. \citet{lenjani2020automated} developed a model from panoramic street view images to extract building images in multiple viewpoints for post-disaster evaluation. \citet{khajwal2023post} proposed a building damage classification model using multi-view feature fusion; however, because the front door is only visible in limited viewpoints in street view images, single-view LFE estimation is required, which increases the difficulty of estimation. The existing single-view LFE estimation method using image segmentation can only provide LFE estimations for approximately 60\% of houses with visible front doors due to challenges such as the quality of segmentation masks and the capability to distinguish front doors from other doors \citep{ho2023elevvision}. A conventional way to improve segmentation performance is to create a high-quality training set for specific objectives. In this case, it would be a panoramic street view image dataset specifically for front doors in outdoor scenes. Nevertheless, generating such datasets can be labor-intensive and sensitive to variations across study areas, necessitating model training or fine-tuning for each task.


With the rapid advancements in vision foundation models and prompt engineering, new possibilities are emerging for tackling the existing image segmentation tasks. The Segment Anything model (SAM) \citep{kirillov2023segment} stands out as the foundation model for image segmentation, capable of segmenting every object in an image and generating high-quality masks. SAM's promptable nature facilitates zero-shot generalization, suggesting progress towards implementing image segmentation without the need for training on the dataset for specific task. SAM accommodates flexible input prompts like points or bounding boxes. However, to realize open-vocabulary image segmentation using SAM, a text encoder is still necessary. To broaden SAM's applicability to open-vocabulary tasks, the integration of vision language models (VLMs), such as CLIP \citep{radford2021learning}, presents a promising avenue. VLMs extract both text and image features, learning image-text correspondences \citep{zhang2024vision}. \citet{li2023clip} proposed CLIP Surgery, enhancing the explainability of CLIP and facilitating its integration with SAM by converting outputs of CLIP Surgery into point prompt inputs for SAM. Grounding DINO \citep{liu2023grounding} outputs have also been utilized as box prompt inputs for SAM \citep{ren2024grounded}. In addition, efforts have been made to integrate Grounding DINO and CLIP with SAM for semantic segmentation in remote-sensing images \citep{zhang2023text2seg}. Despite the emergence of numerous open-vocabulary segmentation models, as comprehensively reviewed by \citet{wu2024towards}, their mask predictions may not match the precision of SAM. Ensuring mask quality is a primary issue to be addressed in LFE estimation task based on street view imagery.

To address the challenge in LFE estimation using image segmentation, our focus lies in conducting text-prompt image segmentation on street view images through the integration of SAM with vision language models. By assessing various vision language models, integration methods, and text prompts with varying levels of detail, we identify the most suitable model for the LFE estimation task on street view images. Leveraging SAM's ability to generate precise masks and the vision language models' capacity to comprehend localized or customized descriptions, our objective is to enhance the estimated LFE on more buildings based on street view imagery with reliable accuracy. Specifically, we aim to increase the proportion of houses for which ELEV-VISION can provide LFE estimations. Moreover, to our best knowledge, this study presents the first baseline and comparison of the performance of different vision models on street view image segmentation for urban analytics tasks. Also, the novel computational model based on integrating vision language and vision foundation models presented in this study can advance vertical feature extraction tasks in civil and infrastructure engineering applications such as detecting structural anomalies in bridges or assessing power infrastructure damage, as well as urban analytics use cases for implementing image segmentation without the need for labor-intensive labeled datasets.

\begin{figure}[ht]
    \centering
    \includegraphics[width=\textwidth]{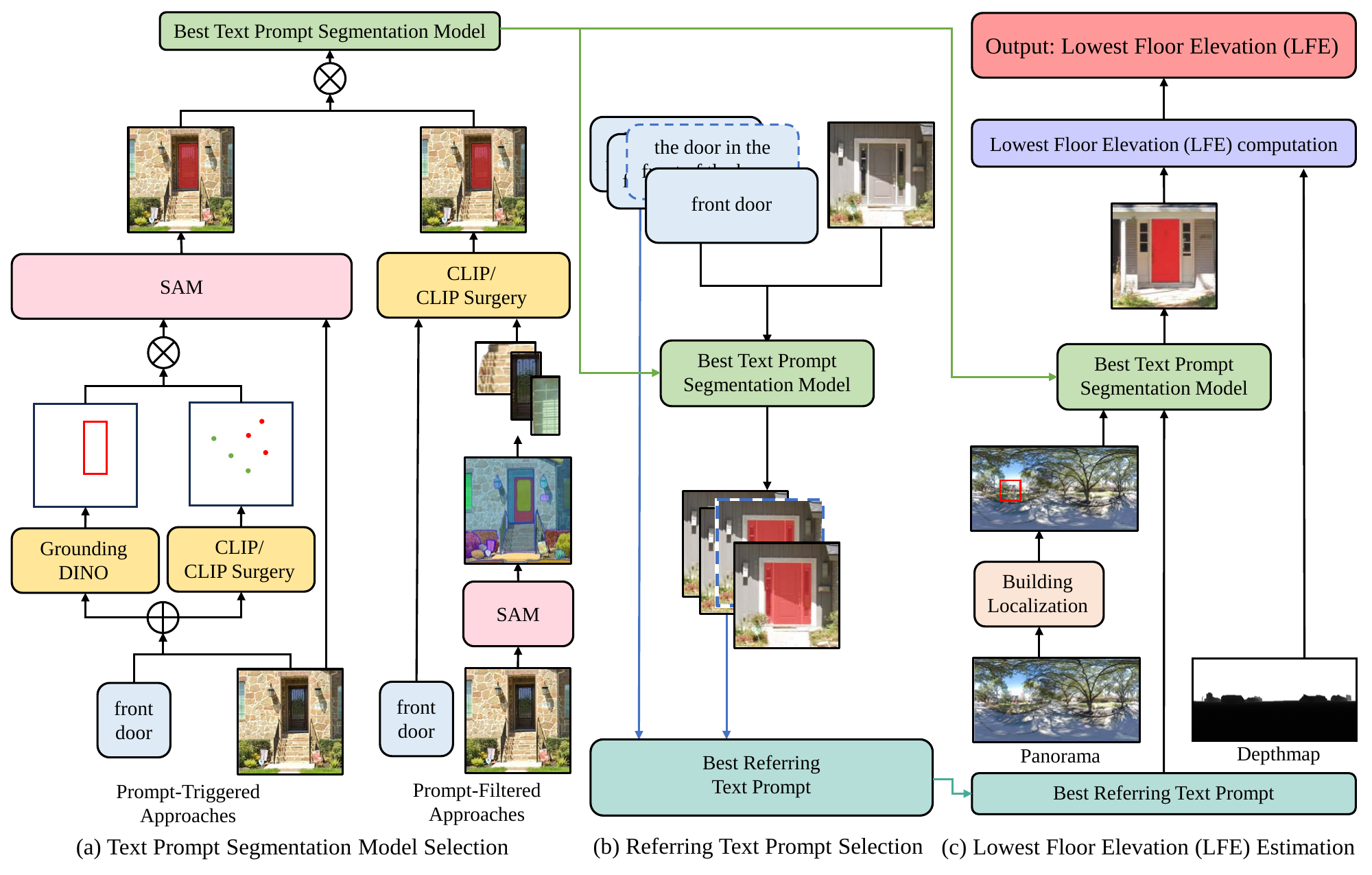}
    \caption{Study workflow. The study consists of three sequential components: text-prompt segmentation model selection, referring text-prompt selection, and LFE estimation. First, a text-prompt segmentation model with the best performance is selected. Next, a referring text prompt is selected to enhance segmentation of the front door of the house. Finally, the selected text-prompt segmentation model and the determined referring text prompt are integrated into the LFE estimation model.}
    \label{fig:conceptual_figure}
\end{figure}

The workflow of this study is depicted in Figure \ref{fig:conceptual_figure}. We utilize vision language models and vision foundation models for text-prompt street view image segmentation to improve LFE estimation. The analysis comprises three sequential components: selection of the text-prompt segmentation model, determination of referring text prompts, and integration with the LFE estimation model.

\section{Methodology}
\label{sec:2}
There are mainly two approaches to text-prompt image segmentation based on SAM: the prompt-triggered approach and the prompt-filtered approach. Both methods operate as two-stage processes, combining visual language models with SAM. In the prompt-triggered approach, a VLM precedes SAM, encoding texts and images to convert text prompts into other prompt types, such as points or boxes, to activate SAM. On the other hand, in the prompt-filtered approach, a VLM succeeds SAM, encoding texts and images to filter the outputs of SAM by image-text similarity.

\subsection{Prompt-triggered Approach}
SAM supports box prompts and point prompts as prompt types, necessitating the conversion of text prompts to either of these formats. Conversion can be directly achieved by VLMs or through their downstream tasks. Integrating CLIP or CLIP Surgery with SAM is the former method. CLIP utilizes contrastive loss to learn text-image similarity, while CLIP Surgery enhances CLIP's explainability by refining self-attention mechanisms and eliminating redundant features. To convert text prompts to point prompts, \citet{li2023clip} used similarity maps generated from CLIP or CLIP Surgery outputs, identifying high-similarity regions to create corresponding points. Alternatively, VLM downstream tasks, such as object detection, offer another avenue. For instance, to convert text prompts to box prompts, open-vocabulary object detection serves as a straightforward solution. Grounding DINO, an open-set detector, extends Transformer-based closed-set detectors by proposing multi-level feature fusion, leveraging similarities between Transformer-based detectors and language models. In this study, we implemented and evaluated three prompt-triggered approaches: CLIP-SAM (SAM triggered by CLIP), CLPS-SAM (SAM triggered by CLIP Surgery), and GDINO-SAM (SAM triggered by Grounding DINO). The implementation of CLIP-SAM and CLPS-SAM follows \citet{li2023clip}'s work; GDINO-SAM implementation is based on the approach proposed by \citet{zhang2023text2seg}. The algorithm of the prompt-triggered approach is shown in Algorithm \ref{alg:trigger}. We choose these VLMs because they represent two different text encoding methods. CLIP-based methods create a sentence for each category, extracting sentence-level features; grounding-based methods concatenate all categories to a string, extracting word-level features.

\begin{algorithm}
\caption{Prompt-triggered approach}\label{alg:trigger}
\SetKwInOut{Input}{Input}
\SetKwInOut{Output}{Output}

\Input{Text prompt $T$, image $I$,\\and vision language model $VLM \in \{"Grounding DINO", "CLIP", "CLIP Surgery"\}$}
\Output{Segmentation masks $M$}
Image features $X_{img} \gets image\_encoder(I)$\;
\uIf{$VLM = "Grounding DINO"$}{
    Text features $X_{text} \gets text\_encoder(T)$\;
    $X_{img}, X_{text} \gets feature\_enhancer(X_{img},X_{text})$\tcp*{feature fusion based on cross-attention}
    \algolines{Cross-modality queries $Q \gets mixed\_query\_selection(X_{img},X_{text})$}{initializing decoder queries by dynamic anchor boxes and static content queries}
    $Q \gets cross\_modality\_decoder(Q)$\tcp*{feature fusion based on cross-attention}
    $Boxes, Logits, Classes \gets anchor\_and\_class\_update(Q)$\;
    Prompts $P \gets Boxes[Classes = "door"]$\; 
  }
\Else{
    Sentences $S\gets sentence\_template(T)$\;
    Text features $X_{text} \gets text\_encoder(S)$\;
    \If{$VLM = "CLIP Surgery"$}{
    Empty string $S_e\gets ""$\;
    Redundant features $X_{r} \gets text\_encoder(S_e)$\;
    $X_{text} \gets X_{text} - X_{r}$\;
      }
    Similarity $Sim \gets cosine\_similarity(X_{img},X_{text})$\;
    Similarity map $Map_{sim} \gets get\_similarity\_map(Sim)$\;
    $Points, Labels \gets map\_to\_points(Map_{sim})$\tcp*{generating foreground and background points}
    Prompts $P \gets [Points, Labels]$;
    }
Segmentation masks $M \gets SAM\_predictor(I,P)$
\end{algorithm}

\subsection{Prompt-filtered Approach}
VLMs can also be integrated with SAM using the prompt-filtered approach. In this method, SAM operates irrelevantly to prompts but rather segments all elements within an image indiscriminately. Specifically, SAM generates a predetermined number of point prompts across the image, followed by the removal of low-quality and duplicate masks. Subsequently, VLMs are employed to filter the segmented masks by establishing correspondences between images and texts. In this study, CLIP and CLIP Surgery are used to identify SAM outputs where the probability of belonging to the "front door" category exceeds a given threshold. We implemented two prompt-filtered approaches: SAM-CLIP (SAM filtered by CLIP) and SAM-CLPS (SAM filtered by CLIP Surgery), adopting the implementation from \citet{segment_anything_with_clip}. The algorithm of the prompt-filtered approach is shown in Algorithm \ref{alg:filter}.

\begin{algorithm}
\caption{Prompt-filter approach}\label{alg:filter}
\SetKwInOut{Input}{Input}
\SetKwInOut{Output}{Output}

\Input{Text prompt $T$, image $I$,\\and vision language model $VLM \in \{"CLIP", "CLIP Surgery"\}$}
\Output{Segmentation masks $M$}
Segmentation masks $M \gets SAM\_generator(I)$\;
Sentences $S\gets sentence\_template(T)$\;
Text features $X_{text} \gets text\_encoder(S)$\;
\If{$VLM = "CLIP Surgery"$}{
Empty string $S_e\gets ""$\;
Redundant features $X_{r} \gets text\_encoder(S_e)$\;
$X_{text} \gets X_{text} - X_{r}$\;
  }
$Labels[0...length(M)] \gets [0...0]$\;
\For{$i = 0;\ i < length(M);\ i = i + 1$}{
    Cropped image $I_{crop} \gets crop\_image(I,M[i])$\;
    Cropped image features $X_{img, crop} \gets image\_encoder(I_{crop})$\;
    Similarity $Sim \gets cosine\_similarity(X_{img},X_{text})$\;
    \If{$Sim>threshold$}{
    $Labels[i] \gets 1$
    } 
  }
$M \gets M[Labels=1]$\;
\end{algorithm}

\subsection{Integration with LFE Estimation}
\begin{figure}[ht]
    \centering
    \includegraphics[width=\textwidth]{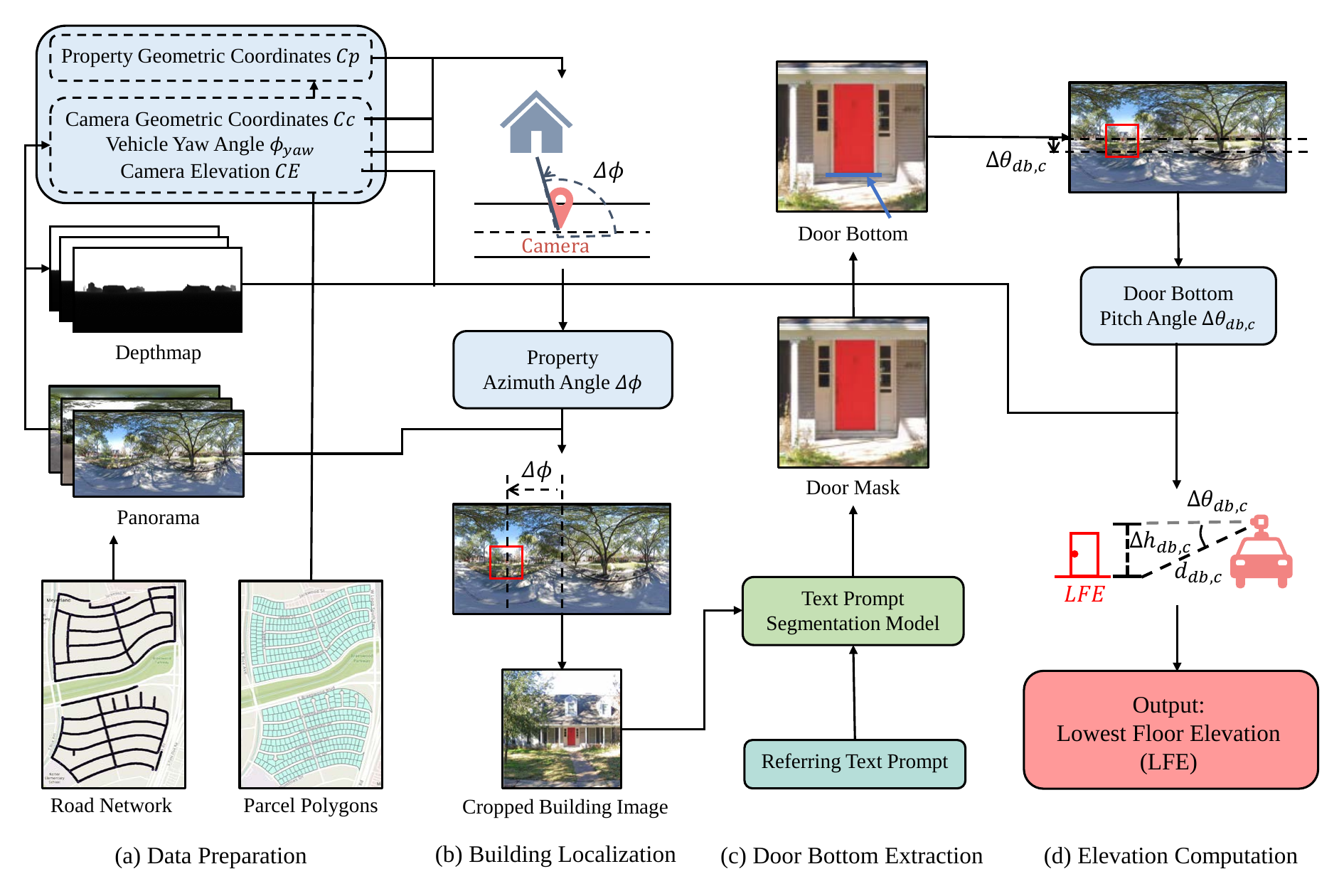}
    \caption{The framework of LFE estimation. The framework consists of four components: data preparation based on road network and parcel data, building localization based on equirectangular projection, door bottom extraction via text-prompt image segmentation, and elevation computation using equirectangular projection, depthmap, and trigonometry.}
    \label{fig:Elev-Vision-SAM}
\end{figure}

In this study, we implemented and compared three prompt-triggered approaches and two prompt-filtered approaches for text prompt street view image segmentation. The approach demonstrating the best performance is integrated with ELEV-VISION, an LFE estimation model based on street view images, to enhance its availability. The proposed LFE estimation model comprises three components: building localization using equirectangular projection principle and camera information, extraction of door bottoms via text-prompt image segmentation, and elevation computation based on equirectangular projection, depthmap, and trigonometry. We replace the image segmentation model in ELEV-VISION with a text-prompt image segmentation method to enhance the segmentation of front doors and the quality of masks. The algorithm of LFE estimation is shown in Algorithm \ref{alg:LFE} and depicted in Figure \ref{fig:Elev-Vision-SAM}.

\begin{algorithm}
\caption{LFE estimation}\label{alg:LFE}
\SetKwInOut{Input}{Input}
\SetKwInOut{Output}{Output}

\Input{Property geometric coordinates $C_p$, panoramic image $I$, panoramic depthmap $D$, camera geometric coordinates $C_c$, camera elevation $CE$, street view vehicle yaw angle $\phi_{yaw}$, text prompt $T$}
\Output{$LFE$}
Property bearing angle from camera $\phi_{p,c} \gets bearing\_angle\_computing(C_c,C_p)$\;
$\Delta\phi \gets \phi_{p,c}-\phi_{yaw}, \Delta\phi \in [-180, 180]$\tcp*{building localization} 
Property image $I_{p} \gets crop\_image(I,\Delta\phi)$\;
Door mask $m \gets text\_prompt\_segmentation(I_{p},T)$\;
Door bottom points $P_{db} \gets door\_bottom\_extraction(m)$\;
Door bottom distances from camera $D_{db,c} \gets D[P_{db}]$\;
Door bottom pitch angles from camera $\Delta\Theta_{db,c} \gets (\frac{height(I)}{2} - y_{db}) \cdot \frac{180}{height(I)}$\;
Door bottom height differences from camera $\Delta H_{db,c} \gets D_{db,c} \cdot sin(\Delta\Theta_{db,c})$\;
Door bottom elevations $DBE \gets CE + \Delta H_{db,c}$\;
$LFE \gets median(DBE)$
\end{algorithm}

\subsubsection{Data Preparation}
The first step is to prepare the input data in Algorithm \ref{alg:LFE}. First, we extracted the road network of the study area from OpenStreetMap \citep{OpenStreetMap} and download Google Street View panoramic images along the roads. Since high-resolution panoramic street view images are not able to be directly downloaded from Google Street View API, we automatically download the tiles of the panoramic images, each with a resolution of 512 × 512 pixels, and concatenate them. The resolution of the concatenated panoramic images is 8192 x 16384 pixels or 6656 x 13312 pixels. The street view image providing the optimal viewpoint to capture the bottom of the front door for each property is then selected. The associated depthmaps and meta-data of the selected images are downloaded. The depthmap represents the distances from the camera to the objects in the image. The original depth information downloaded from Google Street View API is in Base64 format. To pair depthmaps with optical street view images, we decode Base64 strings to depth images. The resolution of the decoded depth images is 256 x 512 pixels. The meta-data used for LFE estimation contains camera geometric coordinates \(C_c\), camera elevation \(CE\), and street view vehicle yaw angle \(\phi_{yaw}\) (relative to North). Thus far, the remaining input data in Algorithm \ref{alg:LFE} is property geometric coordinates \(C_p\). Property geometric coordinates are extracted from the parcel data from City of Houston Geographic Information System \citep{COHPARCELS}.

\subsubsection{Building Localization}
\begin{figure}[ht]
    \centering
    \includegraphics[width=\textwidth]{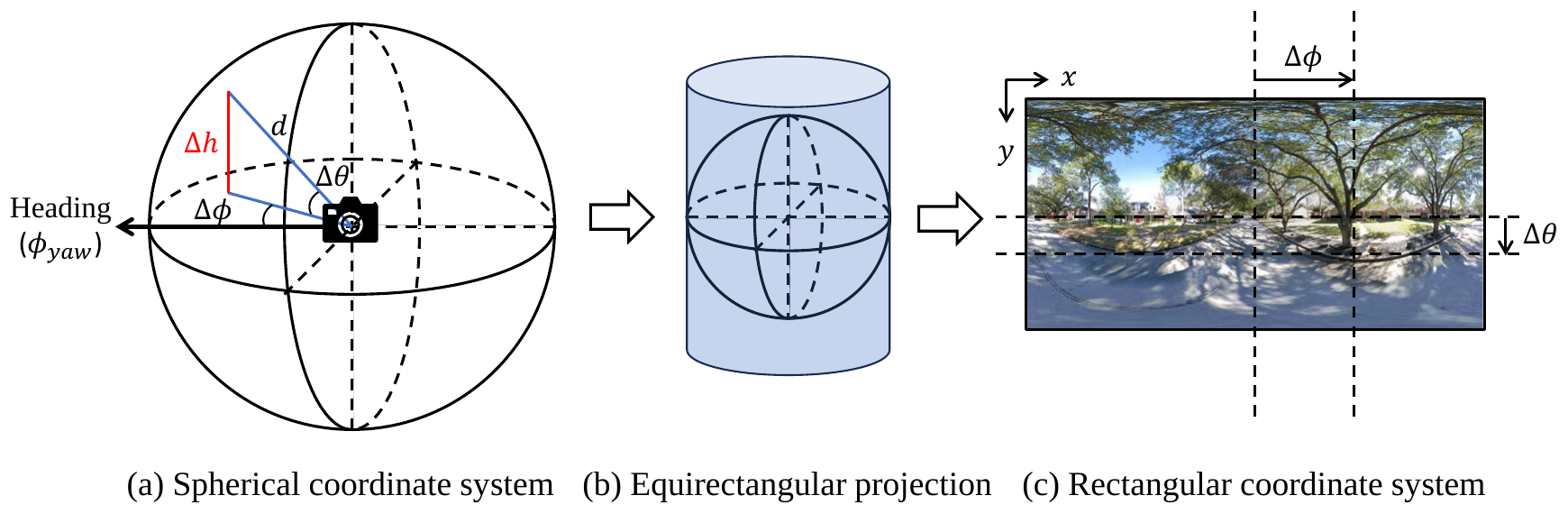}
    \caption{Conversion between spherical and rectangular coordinate system. The center in the spherical coordinate system is the location of the camera. The location of a given point in the spherical coordinate system is represented by \((d, \Delta\theta, \Delta\phi)\), which can be converted to rectangular coordinate representation based on linear spacing of degree difference.}
    \label{fig:spherical_coordinate_system}
\end{figure}

Based on the equirectangular projection principle, we can convert between the spherical coordinate system and the rectangular coordinate system, thereby locating the building in the panoramic image from the azimuth angle of the building relative to the street view vehicle heading direction. Figure \ref{fig:spherical_coordinate_system} depicts the conversion between spherical and rectangular coordinate system. 
First, the bearing angle from the camera to the property \(\phi_{p,c}\) (relative to North) can be calculated from camera geometric coordinates \(C_c (lat_c, lon_c)\) and property geometric coordinates \(C_p (lat_p, lon_p)\), as shown in Eq. \ref{eq:bear}.
\begin{equation}\label{eq:bear}
\phi_{p,c}  = atan2(X, Y) \cdot \frac{180}{\pi}
\end{equation}
\begin{equation}
X = sin(lon_p - lon_c)\cdot cos(lat_p)
\end{equation}
\begin{equation}
Y = cos(lat_c)\cdot sin(lat_p) - sin(lat_c)\cdot cos(lat_p)\cdot cos(lon_p - lon_c)
\end{equation}
\noindent
The azimuth angle of the property \(\Delta\phi\) is the angle difference between the bearing angle from the camera to the property \(\phi_{p,c}\) and the street view vehicle yaw angle \(\phi_{yaw}\), as shown in Eq. \ref{eq:azimuth}.
\begin{equation}\label{eq:azimuth}
\Delta\phi = \phi_{p,c}-\phi_{yaw}, \quad\Delta\phi \in [-180, 180]
\end{equation}
\(\Delta\phi\) is set to be between -180\textdegree and 180\textdegree because the middle of the panoramic image represents the street view vehicle yaw angle \(\phi_{yaw}\). Using \([-180, 180]\), we can better represent the location of the property. With the azimuth angle of the property \(\Delta\phi\) in the spherical coordinate system, we are able to obtain the x value of the property coordinates \(x_p\) in the rectangular coordinate system, as shown in Eq. \ref{eq:loc}.
\begin{equation}\label{eq:loc}
x_p = \frac{W_{img}}{2} + \frac{\Delta\phi}{180} \cdot \frac{W_{img}}{2}
\end{equation}
in which \(W_{img}\) is the width of the panoramic image. Then, we crop the building image \(I_p\) based on the property center coordinates \((x_p,y_p)\). As we only require an approximate building location, we set \(y_p\) to 0, considering that buildings typically do not vertically deviate significantly from 0\textdegree.

\subsubsection{Door Bottom Extraction}
Next, we input the cropped image \(I_p\) and the text prompt \(T\) into the image segmentation model and extract the door bottom from the mask output. The mask \(m\) is a matrix of the same size as the cropped image, containing values of 0 or 1, as shown in Eq. \ref{eq:mask}.
\begin{equation}\label{eq:mask}
m[x,y] = \left\{
\begin{array}{rcl}
1 & & pixel\:(x,y)\:represents\:the\:door\\
0 & & else
\end{array} \right.
\end{equation}
Then, Eq. \ref{eq:door} extracts the door bottom points \(P_{db}\) from the mask \(m\). Specifically, we extract all the columns in the mask \(m\) with at least a value of 1. The column number is \(x_{db}\). In each column \(m[x_{db}]\), the lowest row with a value of 1 is extracted as \(y_{db}\).
\begin{equation}\label{eq:door}
(x_{db}, y_{db}) \in P_{db} \quad\forall\quad [\exists y: m[x_{db},y] = 1 \quad and \quad (y_{db} = \max_{m[x_{db},y] = 1} y \quad\forall\quad x_{db})]
\end{equation}
\noindent

\subsubsection{Elevation Computation}
To compute elevation, the pitch angle and the radial distance in the spherical coordinate system are required. The radial distance is the distance from the camera to the door bottom \(d_{db,c}\), which is extracted from the depthmap \(D\), as shown in Eq. \ref{eq:depth}.
\begin{equation}\label{eq:depth}
d_{db,c} = D[x_{db}, y_{db}]
\end{equation}
Using Eq. \ref{eq:depth}, we extract a list of radial distances \(D_{db,c}\) from a list of door bottom points \(P_{db}\). The pitch angle from the camera to the door bottom \(\Delta\theta_{db,c}\) can be converted from the coordinates of the door bottom \((x_{db}, y_{db})\) in the rectangular coordinate system, as shown in Eq. \ref{eq:pitch}. 
\begin{equation}\label{eq:pitch}
\Delta\theta_{db, c} = (\frac{H_{img}}{2} - y_{db}) \cdot \frac{180}{H_{img}}
\end{equation}
in which \(H_{img}\) is the height of the panoramic image. The equation for vertical conversion (Eq. \ref{eq:pitch}) is slightly different from the equation for horizontal conversion (Eq. \ref{eq:loc}) because the range of pitch angles is \([-90, 90]\) and the range of azimuth angles is \([-180, 180]\). Using Eq. \ref{eq:pitch}, we obtain a list of door bottom pitch angles \(\Delta\Theta_{db,c}\) from a list of door bottom points \(P_{db}\). Then, we compute the height difference between the camera and the door bottom \(\Delta h_{db, c}\) from the pitch angle \(\Delta\theta_{db, c}\) and the radial distance \(d_{db, c}\), as shown in Eq. \ref{eq:height}, and obtain a list of the height differences \(\Delta H_{db, c}\).
\begin{equation}\label{eq:height}
\Delta h_{db, c} = d_{db, c} \cdot sin(\Delta\theta_{db, c})
\end{equation}
A list of the elevations of the front door bottom points \(DBE\) can be derived from the camera elevation \(CE\), as shown in Eq. \ref{eq:dbe}.
\begin{equation}\label{eq:dbe}
DBE = CE + \Delta H_{db, c}
\end{equation}
The last step is to compute LFE. To ensure a robust LFE estimation from partially imprecise door bottom lines, we selected the median of the front door bottom elevations instead of the mean as LFE, as shown in Eq. \ref{eq:lfe}. It should be noted that the outliers in door bottom elevations should be removed before the median is computed. Door masks are not usually rectangles. They can be parallelograms because of different viewpoints or can be distorted because of equirectangular projection. The list of door bottom elevations may contain points not actually along the door bottom but from the long vertical side of the door. The elevations of these points would be higher and should be removed.
\begin{equation}\label{eq:lfe}
LFE = median(DBE)
\end{equation}

\section{Experiments and Results}
\label{sec:3}
In this study, two experiments, text-prompt segmentation model selection and referring text-prompt selection, were conducted to ascertain the optimal configuration of open-vocabulary segmentation for our proposed method, which is called ELEV-VISION-SAM. Subsequently, we evaluated and compared the availability and performance of ELEV-VISION-SAM against those of the baseline model, the ELEV-VISION model.

\subsection{Study Area and Ground Truth Data}
In this study, LFE measurements acquired through unmanned aerial vehicle system-based photogrammetry serve as the ground truth for evaluating the LFE estimates generated by our proposed method. Detailed procedures for these drone-based measurements are elaborated by \citet{diaz_deriving_2022}. Drone-based LFE measurements were adopted due to their closer alignment with the LFE definition used in street view image-based methods, compared to the definition provided in Elevation Certificates \citep{fema_appendix_2020}. The study area, situated in Meyerland within Harris County, Texas, was selected due to its vulnerability to flooding, highlighting the necessity for accurate LFE information in the region. Both the ground truth data and the study area are consistent with those employed in evaluating the baseline ELEV-VISION model, facilitating a direct comparison between our proposed method (ELEV-VISION-SAM) and the existing baseline model.

\subsection{Baseline Model and Dataset}
The baseline model for LFE estimation used to evaluate our proposed method is ELEV-VISION, which directly estimates LFE from panoramic street view images. In this study, we employed Google Street View images for LFE estimation. Google Street View service was selected for this study due to its stable image quality, consistent acquisition method, comprehensive associated information, and extensive area coverage. The acquisition of street view images is highly standardized within the Google Street View service, with nearly all images offering 360-degree coverage. Moreover, these street view images are accompanied by depth information and various image and camera details such as capture date, location, and camera elevation. The aforementioned benefits facilitate data processing and LFE computation. The dataset description is provided in Table \ref{tab:data}, with an effective data size of 409 building images. Within these 409 buildings, 232 (56.72\%) have visible front doors in the street view images, including those previously identified as visible in the Elev-Vision baseline model \citep{ho2023elevvision} and those obscured by railings. This subset of 232 houses constituted the test set for evaluating LFE estimation. In addition, we assembled two validation sets, comprising images from houses in Meyerland and Edgebrook, another flood-prone neighborhood in Harris County, Texas. The proportions of the two validation sets and the test set are 15\%, 15\%, and 70\%, respectively. LabelMe \citep{labelme}, an open-source image annotation tool written in Python, is used to label the front doors in panoramic images to build the datasets.

\begingroup

\setlength{\tabcolsep}{10pt} 
\renewcommand{\arraystretch}{1.5} 

\begin{table}[htb]
\caption{Description of Data Size.}
\centering
\begin{tabular}{cc}
\cline{1-2}
Data Description & Number of Houses \\
\cline{1-2}
Houses with LFE ground truth and an SVI & 409\\
Houses with a visible front door in the SVI      & 232\\
Houses with the detected door bottom in the SVI\textsuperscript{*} & 229              \\
\cline{1-2}
\multicolumn{2}{l}{\textsuperscript{*} Detailed explanation is provided in Section \ref{sec:3.6}}.
\end{tabular}
\label{tab:data}
\end{table}
\endgroup

\subsection{Evaluation Metrics}
The most common evaluation metrics for segmentation are Intersection over Union (IoU) and Average Precision (AP). For segmentation model selection, IoU is utilized as it better reflects mask quality compared to AP. IoU is calculated as the intersection of predicted masks and ground truth masks divided by their union. Specifically, IoU is represented as
\begin{equation}\label{eq:IoU}
IoU = \frac{Intersection}{Union} = \frac{TP}{TP+FP+FN}
\end{equation}
where \(TP\) denotes the number of front door pixels correctly classified as front door, \(FP\) represents the number of background pixels mistakenly classified as front door, and \(FN\) indicates the number of front door pixels incorrectly classified as background. Additionally, frames per second (FPS) is considered to compare the inference time of models, which is another crucial evaluation metric for segmentation model selection. For text-prompt selection, AP\(_{50}\) is used to assess the ability to segment the objects of interest. In this experiment, we focus on correctly identifying objects rather than precisely delineating mask boundaries, as the accuracy of mask boundary would be similar using the same segmentation model. AP is defined as the mean precision at different recall levels, which is irrelevant to confidence threshold selection. AP\(_{50}\) means the threshold to determine \(TP\) is IoU 50\%. Computation of AP follows the interpolated method used in VOC2010 \citep{pascal-voc-2010}:
\begin{equation}\label{eq:AP}
AP=\sum_{k=0}^{n-1}(r_{k+1}-r_k)P_{interp}(r_{k+1})
\end{equation}
where \(r_k\) is the \(k^{th}\) recall level. The interpolated precision \(P_{interp}(r)\) at recall level \(r\) is the maximum measured precision for which \(\tilde{r}\) is larger or equal to \(r\):
\begin{equation}\label{eq:P_interp} 
P_{interp}(r)=\max_{\tilde{r}:\tilde{r}\geq r} p(\tilde{r})
\end{equation}
For LFE measurement, we employ mean absolute error (MAE) and availability rate as evaluation metrics to assess the performance of our proposed method. MAE is selected as it effectively reflects the impact of LFE measurement accuracy on flood risk assessment. Additionally, availability rate, defined as the percentage of properties for which our model can provide LFE estimation out of the total number of properties, serves as a key metric aligning with the study objective.

\subsection{Text-prompt Segmentation Model Selection}
The first step is to determine the best method to use as the open-vocabulary segmentation model for segmenting front doors. We evaluated five methods, CLIP-SAM, CLPS-SAM, GDINO-SAM, SAM-CLIP, and SAM-CLPS, on our first validation set. To ensure comparability, we employed a ViT-B \citep{dosovitskiy2020image} backbone for CLIP and CLIP Surgery, while a Swin-B \citep{liu2021swin} backbone was chosen for Grounding DINO. The text prompt used in this experiment was \textit{front door}. The IoU (\%) and FPS of each configuration are presented in Table \ref{tab:seg_result}. The inference time is measured on RTX A6000.

\begingroup

\setlength{\tabcolsep}{10pt} 
\renewcommand{\arraystretch}{1.5} 

\begin{table}[htb]
\caption{Results of five segmentation methods.}
\centering
\begin{tabular}{ccccc}
\cline{1-5}
Model     & VLM Backbone & SAM Backbone & IoU (\%) & FPS  \\
\cline{1-5}
CLIP-SAM  & ViT-B        & ViT-H        & 6.64     & 0.9  \\
CLPS-SAM & ViT-B        & ViT-H        & 25.05    & 0.64 \\
GDINO-SAM & Swin-B       & ViT-H        & \textbf{75.63}    & \textbf{1.33} \\
SAM-CLIP  & ViT-B        & ViT-H        & 39.21    & 0.22 \\
SAM-CLPS  & ViT-B        & ViT-H        & 42.70    & 0.10
\\
\cline{1-5}

\end{tabular}
\label{tab:seg_result}
\end{table}
\endgroup

GDINO-SAM achieves the highest IoU and FPS, signifying its ability to  provide more accurate masks in less inference time. Low IoU of CLIP-SAM suggests that CLIP is not capable of generating appropriate point prompts from text prompts based on its similarity map. The problem is that the similarity map of CLIP is visually opposite owing to the self-attention computed from query and key and noisy because of redundant features \citep{li2023clip}. Although CLIP fails to be integrated with SAM by a prompt-triggered approach, it can be used to filter the outputs of SAM, as evidenced by the higher IoU of SAM-CLIP. SAM-CLIP and SAM-CLPS yield superior IoU compared to CLPS-SAM. This output implies that filtering SAM's outputs might be more precise than triggering SAM through point prompts, albeit at a cost of efficiency. The low FPS of SAM-CLIP and SAM-CLPS is anticipated since both stages of the prompt-filtered approach operate on the entire image, whereas only the first stage of the prompt-triggered approach is processed on the entire image. An anatomy of the slowest method, SAM-CLPS, reveals that SAM consumes an average of 2.34 seconds to generate all masks, while CLIP Surgery requires 6.09 seconds on average to compute similarities between objects and texts. Notably, the SAM step alone in SAM-CLPS took even longer time than the total processing time of any prompt-triggered approach. In addition, the potential presence of numerous overlapping masks might amplify the image size processed by CLIP Surgery, thereby prolonging its runtime.

\begin{figure}[p]
    \centering
    \includegraphics[width=0.9\textwidth]{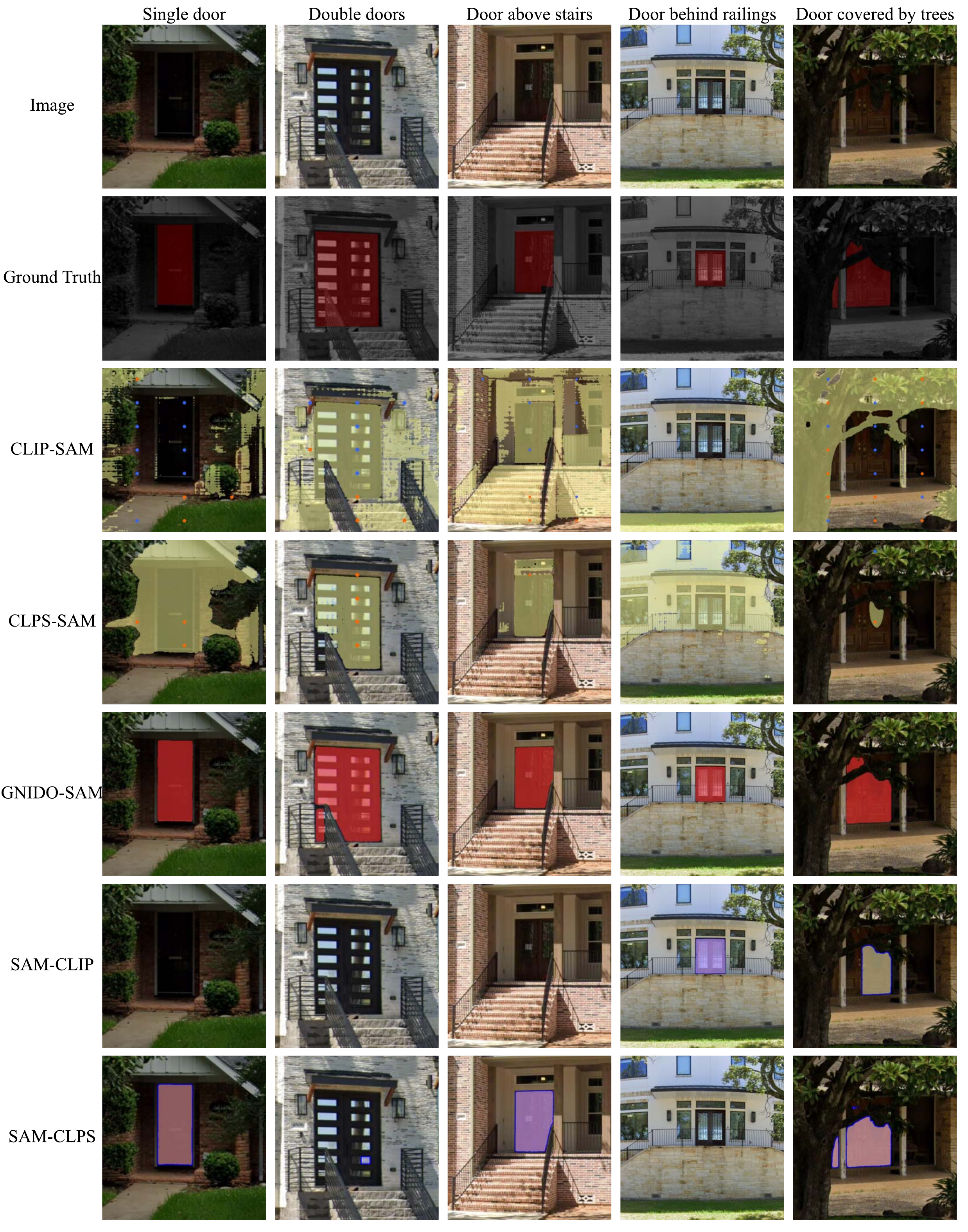}
    \caption{Performance visualization of the five segmentation methods for various types of doors. GDINO-SAM stands out as the most accurate method overall, effective in identifying small setions of doors and performing well in obstructed scenarios, such as those concealed behind railings or trees, and in low light conditions. CLPS-SAM, SAM-CLIP, and SAM-CLPS exhibit limited success in segmenting doors for specific scenarios. Notably, SAM-CLPS outperforms SAM-CLIP in low light conditions. CLIP-SAM fails to generate masks for any doors.}
    \label{fig:model_selection_examples}
\end{figure}

The performance of the five methods in different scenarios is depicted in Figure \ref{fig:model_selection_examples}. GDINO-SAM consistently produced the most accurate masks overall. For the single door example, only GDINO-SAM and SAM-CLPS were successful in generating the mask, suggesting potential difficulty due to low-light conditions. In the case of double doors, GDINO-SAM generated the most complete mask while CLPS-SAM was also able to generate the mask, suggesting potential challenges arising from their window-like appearance or partial obstruction by railings. In scenarios involving doors above stairs, GDINO-SAM and SAM-CLPS exhibited the most precise segmentation, while CLPS-SAM also performed adequately, suggesting potential challenges associated with recessed positioning, thereby lying in low light condition, and occlusion by handrails. When dealing with doors behind railings, only GDINO-SAM and SAM-CLIP were capable of generating masks, indicative of the inherent difficulty in segmenting such doors even under decent lighting conditions. For doors covered by trees, GDINO-SAM and SAM-CLPS yielded more accurate masks. GDINO-SAM excelled in capturing small pieces of the door amidst foliage. SAM-CLPS segmented the small part of the door between the tree trunk and the pole. In contrast, SAM-CLIP struggled to segment the entire door, managing only partial success. Overall, GDINO-SAM exhibits robust performance across various scenarios, effectively identifying door features even in challenging conditions such as low light and obstruction by railings or trees. CLPS-SAM, SAM-CLIP, and SAM-CLPS exhibit partial success in door segmentation across certain instances. Notably, SAM-CLPS outperforms SAM-CLIP in low light conditions.

\subsection{Referring Text-prompt Selection}
After selecting GDINO-SAM based on the previous performance comparison, we encountered some challenges during the experiment. Specifically, distinguishing front doors from other types of doors, such as garage doors or the front doors of cars, was difficult. To address this ambiguity and enhance performance, we explored more detailed descriptions to mitigate such issues. Grounding DINO's capability in referring expression comprehension enables it to distinguish the object to which the user refers from others in the same category. Leveraging this method, we tested five different text prompts on our second validation set, as outlined in Table \ref{tab:text_result}. \textit{Front door} is the text prompt used in the previous experiment. \textit{Door} is used to compare with front door to see if not specifying front door would make a difference. \textit{The door in the front of the house} is designed to emphasize that we are not interested in the front door of other objects such as the front door of the car. \textit{The door for humans in the front of the house} and \textit{the door not for cars in the front of the house} are designed to avoid targeting garage doors.

\begingroup
\setlength{\tabcolsep}{10pt} 
\renewcommand{\arraystretch}{1.5} 

\begin{table}[htb]
\caption{Results of five different text prompts for GDINO-SAM.}
\centering
\begin{tabular}{p{0.39\textwidth}>{\centering} p{0.1\textwidth} p{0.38\textwidth}}
\cline{1-3}
Text Prompt                                       & AP\(_{50}\) (\%) & Meaning of text prompt design                                                                \\
\cline{1-3}
Door                                              & 48.45     & A more general prompt                                                                        \\
Front door                                        & 59.32     & Default prompt in model selection                                                            \\
The door in the front of the house                & \textbf{78.45}     & A more specific prompt to describe front door to avoid selecting the front door of the car \\
The door for humans in the front of the house   & 76.97     & A prompt to distinguish the front door and the garage door                                   \\
The door not for cars in the front of the house & 69.38     & A prompt to distinguish the front door and the garage door          \\                        
\cline{1-3}
\end{tabular}
\label{tab:text_result}
\end{table}
\endgroup

The results of different text prompts are presented in Table \ref{tab:text_result}.  \textit{The door in the front of the house} outperforms other text prompts. We set the result of \textit{Front door} as the baseline. Not specifying front door decreases AP\(_{50}\) by 10.87\%. Emphasizing the front door of the house greatly improves AP\(_{50}\) by 19.13\%. \textit{The door for humans in the front of the house} and \textit{the door not for cars in the front of the house} fail to boost the performance further compared to \textit{The door in the front of the house}. Even the text prompt with the highest AP\(_{50}\) cannot perfectly distinguish front doors and garage doors. Sometimes it selected both the front door and the garage door or it selected only the garage door.

\begin{figure}[p]
    \centering
    \includegraphics[width=0.9\textwidth]{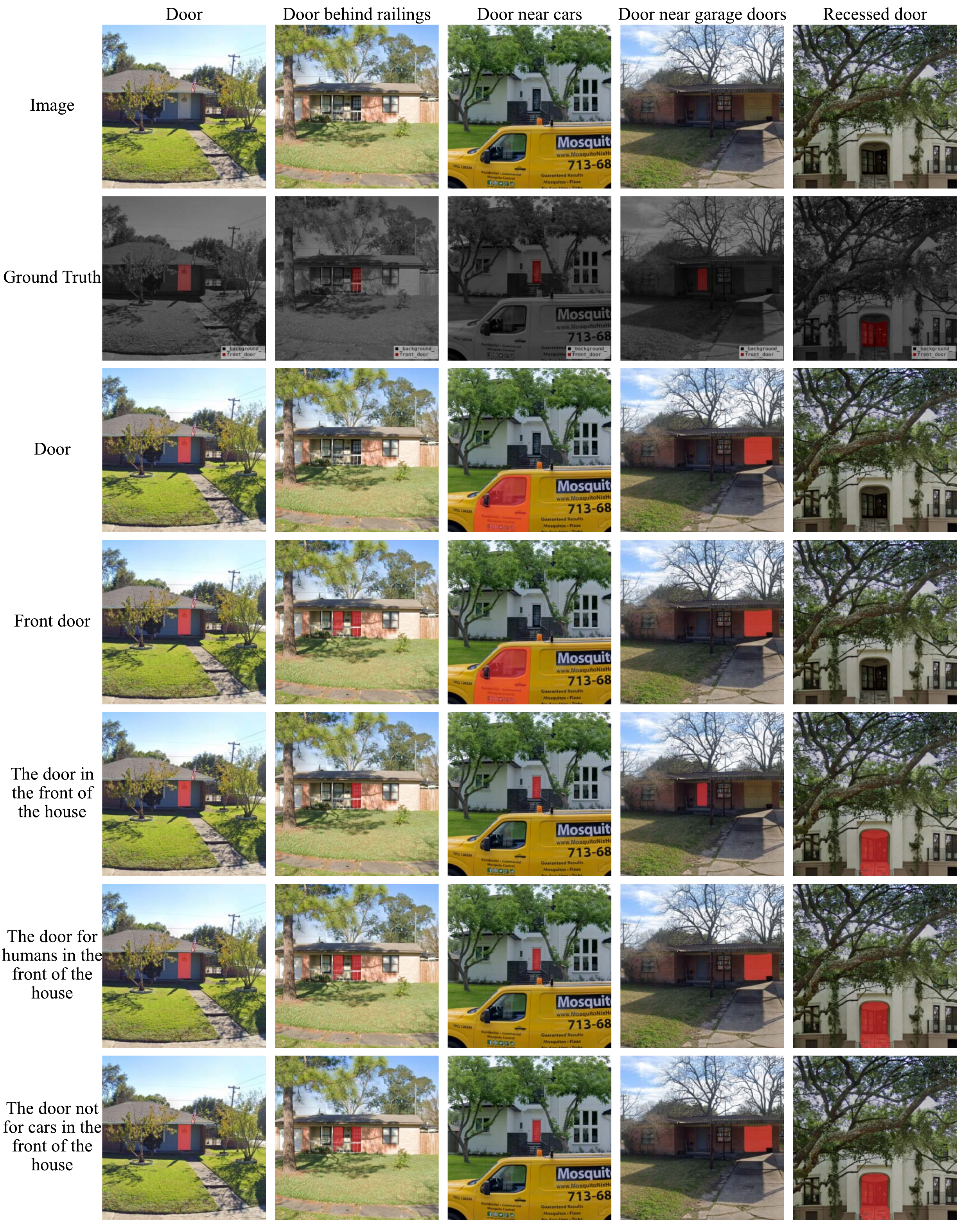}
    \caption{Performance visualization of the five different text prompts for various types of doors. \textit{The door in the front of the house} demonstrates strong performance across most instances, which is the most appropriate text prompt for distinguishing the house's front door from both the front door of a car and the garage door.}
    \label{fig:text_prompt_selection_examples}
\end{figure}

Figure \ref{fig:text_prompt_selection_examples} shows the performance of five text prompt in different cases. For the door not occluded and not close to cars and garage doors, all text prompts could segment the door. For the door behind railings, \textit{the door in the front of the house} generated the most accurate mask, whereas \textit{front door}, \textit{the door for humans in the front of the house}, and \textit{the door not for cars in the front of the house} were able to segment the front door but mistakenly segment the window; \textit{door} failed to segment the front door. For the door near cars, \textit{the door in the front of the house}, \textit{the door for humans in the front of the house}, and \textit{the door not for cars in the front of the house} were able to segment the front door of the house while \textit{door} and \textit{front door} segmented the front door of the car. For the door near garage doors, \textit{the door in the front of the house} was the only text prompt which successfully segmented the front door, whereas other text prompts segmented the garage door. It should be noticed that even \textit{the door in the front of the house} did not always correctly segment the front door. For the door deeply recessed, none of the text prompts demonstrated accurate segmentation of the front door. In addition to the low light condition in light of deep recess, the window-alike appearance and the unusual design could be the difficulties to segment this door. \textit{The door in the front of the house}, \textit{the door for humans in the front of the house}, and \textit{the door not for cars in the front of the house} segmented the arched doorway and stairs together. \textit{Door} and \textit{front door} did not generate any masks. Overall, \textit{the door in the front of the house} exhibited good performance across the majority of instances. It stands out as the most suitable text prompt for distinguishing the front door of the house from both the front door of the car and the garage door. It surpasses the ability of the other two text prompts specifically designed for distinguishing the front door from the garage door.

\subsection{LFE Estimation Performance}
\label{sec:3.6}
Based on the outcomes of the previous two experiments, we have decided to use GDINO-SAM as the segmentation method alongside the text prompt \textit{The door in the front of the house} for front door segmentation in LFE computation. Results and a comparison with the baseline ELEV-VISION model are summarized in Table \ref{tab:LFE_result}. Notably, our proposed method significantly enhances availability of houses for LFE estimation, providing estimations for approximately 56\% of houses (229 houses out of 409). Impressively, our method demonstrates applicability to 98.71\% of houses where the front door is visible, covering 229 out of 232 such houses. Although the mean absolute error of our proposed method was slightly higher than that of ELEV-VISION, both models demonstrated comparable performance when they were applied to only the houses within only ELEV-VISION's available scope. These findings suggest that our approach enhances availability without sacrificing accuracy. In addition, it is worth noting that while ELEV-VISION requires the labeling of the proportion of the detected door bottoms to improve LFE computation, our proposed method does not necessitate this step, as GDINO-SAM can extract precise door bottom information automatically.

\begingroup

\setlength{\tabcolsep}{10pt} 
\renewcommand{\arraystretch}{1.5} 

\begin{table}[htb]
\caption{Results of LFE estimation.}
\centering
\begin{tabular}{cccc}
\cline{1-4}
Model                  & MAE (m) & Availability (\%) & Availability in the houses with visible front doors (\%) \\
\cline{1-4}
ELEV-VISION-SAM (Ours) & 0.22    & \textbf{55.99}             & \textbf{98.71} \\
ELEV-VISION            & \textbf{0.19}    & 33.25             & 58.62                    
\\
\cline{1-4}

\end{tabular}
\label{tab:LFE_result}
\end{table}
\endgroup

\begin{figure}[ht]
    \centering
    \includegraphics[width=\textwidth]{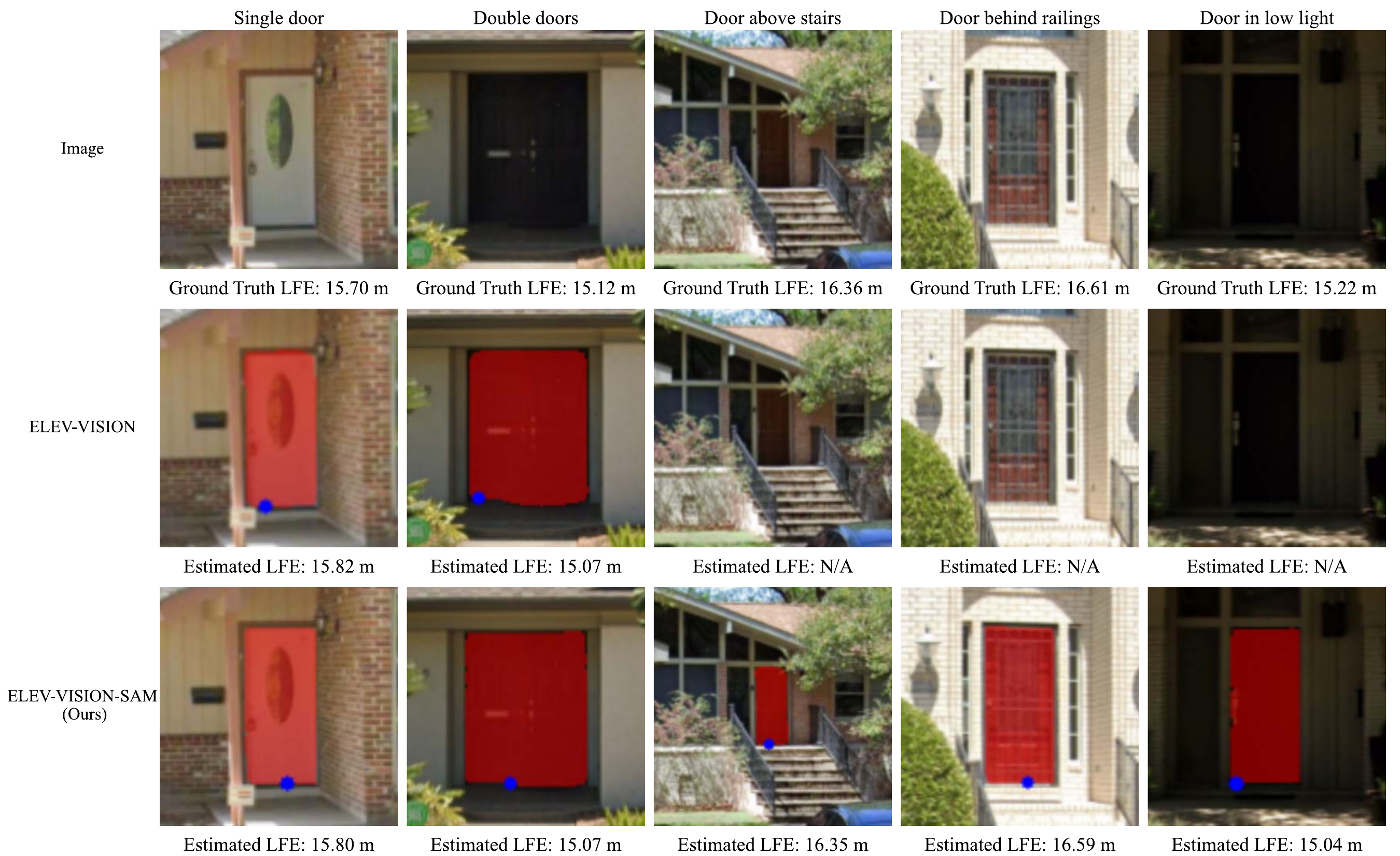}
    \caption{Examples of LFE results. The blue points are the selected points to denote door bottom, with their elevations serving as the estimated LFEs. The LFE results obtained from two methods exhibit similarity in single-door and double-doors cases. However, challenges arise for ELEV-VISION when dealing with doors recessed above stairs, doors obscured by railings, and doors in low light conditions. In contrast, our proposed method demonstrates the capability to provide reliable LFE results under these challenging conditions.}
    \label{fig:LFE_estimation_examples}
\end{figure}

Figure \ref{fig:LFE_estimation_examples} illustrates examples of LFE performance results. The blue points depicted on the door bottoms are selected from a list of door bottom points to represent the door bottom. Specifically, the median of the elevation values is selected as the estimated LFE after outliers are filtered out. In both single-door and double-door scenarios, our proposed method achieves comparable LFE estimation to that of ELEV-VISION. However, our method demonstrates greater capability in providing reliable LFE estimation for challenging cases, such as recessed doors, doors obscured by railings, or doors in low light conditions, where ELEV-VISION struggles.

\section{Concluding Remarks}
\label{sec:4}
The results from the segmentation model selection phase show GDINO-SAM as the optimal choice for segmenting front doors in street view images. Utilizing the bounding box outputs of Grounding DINO as inputs for SAM, GDINO-SAM achieves an Intersection over Union of 75.63\%, outperforming the next-best model by 32.93\%. Notably, GDINO-SAM excels in challenging conditions, such as occlusion and low light, demonstrating robustness across various scenarios. Furthermore, GDINO-SAM exhibits remarkable efficiency, operating at a frame rate of 1.33 frames per second, which represents a more than twofold improvement over the second most efficient model. In addition, the results suggest that approaches triggering SAM by a vision language model yield greater efficiency compared to those filtering SAM's outputs by a VLM. Specifically, the prompt-triggered approach demonstrates a speed advantage ranging from four to six times compared to the prompt-filtered approach using the same VLM.

When employing \textit{front door} as the text prompt input for GDINO-SAM, the model encountered challenges in accurately discerning the front door of the house amidst other doors. To enhance the model's capability in targeting the front door, a more specific text prompt is required. Results from text-prompt selection indicate that \textit{the door in the front of the house} stands out as the most effective text prompt in distinguishing the front door of the house from both car doors and garage doors, achieving an AP\(_{50}\) of 78.45\%. However, limitations persist in precisely classifying front doors and garage doors, even with the use of an optimized text prompt.

Having finalized our text-prompt segmentation model by determining the most suitable segmentation model and text prompt, we integrated it into the LFE estimation model. Our proposed model significantly enhances the availability of LFE estimation, achieving an availability rate of 56\%, outperforming the state-of-the-art model (ELEV-VISION) by 22.74\%. Notably, our model can estimate LFE for nearly all houses with visible front doors, boosting ELEV-VISION's availability to houses with visible front doors by 40.09\%. Importantly, this enhancement in availability does not compromise reliability, as our model achieves a comparable MAE with ELEV-VISION. However, challenges persist in further improving the MAE, as evidenced by instances where more precise segmentation masks did not lead to enhanced LFE estimation. This finding suggests that limitations in depthmap resolution may hinder accurate estimation, causing different positions to correspond to the same depth due to low resolution.

This study contributes to enhancing automated estimation of the lowest floor elevation of buildings by employing vision language and foundation models on street view imagery. The main computational contribution of this study is that it presents the first comprehensive comparison of various approaches using vision language and vision foundation models to text-prompt image segmentation on street view images and the results show significant improvement of the availability of the existing LFE estimation model. In this study, we evaluate the effectiveness and the efficiency of implementing text-prompt segmentation with different vision language models, different integrated structure, and different text prompts. The study identifies integrating Grounding DINO, an open-set object detector, and SAM, a segmentation foundation model, as the optimal text-prompt segmentation model for accurately segmenting front doors in street view images, achieving significant dominance in both performance and efficiency compared to alternative models, especially in challenging conditions such as occlusion and low light. Additionally, the investigation into text-prompt selection provided valuable insights into the importance of using specific referring prompts to enhance model performance and reliability. By leveraging these advanced computational techniques, the proposed LFE estimation model outperforms the baseline model in both the availability and efficiency with the comparable error rate.

The outcomes of this study are crucial as they address the pressing need for accurate and efficient LFE estimation, which is essential for effective flood risk prediction and damage estimation. By providing a method to automatically estimate LFE from street view images, the study significantly enhances the availability and reliability of LFE estimation compared with the existing model, thereby potentially improving the overall resilience of communities to flood events. Moreover, the findings advance the state of the art by demonstrating the effectiveness of vision language models and vision foundation models for text-prompt segmentation in the context of vertical information extraction from street view images, paving the way for future research in this area.

Most of the existing literature related to vertical information extraction from street view images either rely on a reference height or utilize multiple images to enhance height computation. A reference height can be measured for fixed-scene tasks but it is difficult to obtain and needs to be assumed for moving-scene tasks, therefore is less reliable. Multi-view height computation is commonly used for structure height estimation but less useful for LFE estimation because of limited visible viewpoints of the front door. The study presents the novel computational innovations on single-view vertical information extraction without an assumed reference height using text-prompt segmentation, equirectangular projection principle, and the depth information associated to panoramic street view images.

The computational methodology presented in this study has broader applications beyond LFE estimation and flood risk assessment, particularly in the civil and infrastructure engineering field. There are a variety of vertical feature extraction tasks with similar challenges or characteristics as LFE estimation. For example, single-view vertical feature extraction on street view images can be applied to structural or mechanical anomaly detection in bridges when the anomalies are visible only in limited viewpoints. Another highly suitable use case for vertical feature extraction from street view images is electrical infrastructure anomaly or damage assessment, such as measuring power line sag or assessing pole condition. Leveraging a series of historical street view images, subsidence in properties can also be assessed. In addition, our presented baseline and comparison for text-prompt image segmentation on street view images also benefit other civil engineering problems requiring image segmentation. For example, text-prompt segmentation enables complicated scene understanding, which can improve architectural design space interpretation or construction site safety management.

To further advance the computational method presented in this study and its implementation, future work could focus on developing enhanced methods to differentiate between front doors and garage doors. Also, novel techniques for extracting depth information directly from street view images, eliminating the reliance on low-resolution depthmaps, could be explored. Moreover, leveraging text prompt segmentation models opens avenues for incorporating additional building features into LFE estimation to further enhance the performance and availability, presenting opportunities for further investigation and refinement in this domain. Finally, future studies could implement the presented method in other vertical feature extraction tasks, such as structurally anomaly detection and comparison of the results with the state-of-the-art models.


\section{Data Availability}
The data that support the findings of this study are available from Google Street View data.

\section{Code Availability}
The code that supports the findings of this study is available from the corresponding author upon request.

\section{Acknowledgements}
We would like to thank Dr. Samuel D. Brody and his Ph.D. student, Nicholas D. Diaz (Texas A\&M University at Galveston), for providing invaluable drone-based data for model evaluation. We thank the undergraduate researcher Andrew Zheng (Texas A\&M University) for help with image annotation. The authors would like to acknowledge funding support from  the National Science Foundation under CRISP 2.0 Type 2,  grant 1832662,  and the Texas A\&M X-Grant Presidential Excellence Fund. Any opinions, findings, conclusions, or recommendations expressed in this research are those of the authors and do not necessarily reflect the view of the funding agencies.

\bibliography{ref}

\end{document}